\UseRawInputEncoding
\documentclass[twocolumn]{article}
\usepackage{cite}
\usepackage{amsmath,amssymb,amsfonts}
\usepackage{algorithmic}
\usepackage{graphicx}
\usepackage{textcomp}

\usepackage{booktabs}
\usepackage{multirow}
\usepackage{subcaption}
\usepackage{authblk} 
\usepackage{float}
\usepackage{hyperref}
\hypersetup{
    colorlinks=true,
    linkcolor=blue,
    citecolor=blue,
    filecolor=magenta,
    urlcolor=cyan
}

\begin{document}
\title{MInDI-3D: Iterative Deep Learning in 3D for Sparse-view Cone Beam Computed Tomography} 

\author[1]{Daniel Barco}
\author[1]{Marc Stadelmann}
\author[1]{Martin Oswald}
\author[2]{Ivo Herzig}
\author[2]{Lukas Lichtensteiger}
\author[3]{Pascal Paysan}
\author[3]{Igor Peterlik}
\author[3]{Michal Walczak}
\author[4]{Bjoern Menze}
\author[1]{Frank-Peter Schilling}

\affil[1]{Centre for Artificial Intelligence (CAI), Zurich University of Applied Sciences (ZHAW), Winterthur, Switzerland.}
\affil[2]{Institute of Applied Mathematics and Physics (IAMP), Zurich University of Applied Sciences (ZHAW), Winterthur, Switzerland.}
\affil[3]{Varian Medical Systems Imaging Lab, Baden, Switzerland.}
\affil[4]{Biomedical Image Analysis and Machine Learning, University of Zurich, Zurich, Switzerland.}

\twocolumn[
  \begin{@twocolumnfalse}
    \maketitle
    \begin{abstract}
    
    Reducing patient radiation exposure in Cone Beam Computed Tomography (CBCT) by acquiring fewer projections introduces severe image artefacts, limiting its clinical utility. To address this challenge, we propose \textbf{MInDI-3D} (\textbf{M}edical \textbf{In}version by \textbf{D}irect \textbf{I}teration in \textbf{3D}), a 3D conditional diffusion framework that restores volumetric data from sparse-view inputs. Our work provides two key contributions: 1) The MInDI-3D model, the first adaptation of the iterative inversion principle to fully 3D medical volumes, which offers a unique, tuneable trade-off between perceptual quality and quantitative fidelity by adjusting the number of inference steps. 2) A new, publicly available, large-scale dataset of 16,182 pseudo-CBCT volumes to facilitate robust training and future research. 
    On an independent real-world CBCT test set, MInDI-3D achieves performance competitive with state-of-the-art methods, yielding a 10.0 dB PSNR gain over standard reconstructions from only 25 projections. This result enables a 16-fold reduction in radiation exposure and demonstrates robust generalization to a new scanner geometry not seen during training. Beyond standard metrics, MInDI-3D reconstructions preserved high anatomical integrity, enabling accurate automated segmentation in task-based evaluations. In a clinical evaluation by 11 radiotherapy professionals, the reconstructions were rated as sufficient for patient positioning across all tested anatomical sites (abdomen, breast, and lung) and were noted to preserve lung tumour boundaries well. 
    
    \end{abstract}
    \vspace{1em} 
  \end{@twocolumnfalse}
]

\section{Introduction}
\label{sec:introduction}
Reconstructing high-quality medical images from sparsely sampled or partial measurements are essential for advancing clinical imaging modalities such as computed tomography (CT), positron emission tomography (PET), and magnetic resonance imaging (MRI). These advancements aim to reduce scan times and patient radiation exposure. Among these modalities, cone beam computed tomography (CBCT) exemplifies both the promise and challenges of sparse sampling.

CBCT is widely used to acquire volumetric X-ray images on radiation therapy treatment devices, such as linear accelerators, in image-guided radiation therapy \cite{jaffray_flat-panel_2002}. It is also employed in interventional radiology, offering high spatial resolution and short scan durations \cite{yoon_initial_2020}. While pre-treatment planning CT offers higher image resolution for the intervention planning, the image of the day is acquired using on-device CBCT. These CBCT scans can enable tumour and organs at risk contouring, dose calculation, ART (adaptive radiation therapy) workflows and precise patient positioning \cite{lavrova_adaptive_2023}. Its clinical use faces the following challenges: First, image quality is often degraded by artefacts from patient motion, metal implants, and undersampled projections \cite{amirian_artifact_2024}. In addition, repeated daily scans over extended treatment periods (up to 40 sessions) raise concerns about cumulative radiation exposure to the patient.

To address the challenges of cumulative radiation exposure, reducing the number of projections, i.e. sparse-view CBCT, has been proposed. Sparse-view CBCT reconstruction, however, introduces streak artefacts -- due to the Nyquist–Shannon sampling theorem being violated -- which degrade image quality and hinder clinical utility. Deep learning-based approaches have emerged as promising solutions to address these challenges, offering the potential to reconstruct high-quality images from limited projection data.

Deep learning models, particularly convolutional neural networks (CNNs) like the U-Net~\cite{ronneberger_u-net_2015}, are well-suited for medical image reconstruction~\cite{zhai_comprehensive_2023}. The U-Net's encoder-decoder architecture has inspired many variants, including the 3D U-Net for volumetric CBCT data. More recently, generative models such as GANs~\cite{goodfellow_generative_2014}, VAEs~\cite{kingma_auto-encoding_2022}, and diffusion models~\cite{ho_denoising_2020} have been extended to conditional image-to-image tasks relevant to medical imaging, including artefact removal, domain translation, and denoising~\cite{ali_generative_2024}.

Diffusion models have significantly advanced image synthesis and restoration, surpassing traditional GANs in conditional and unconditional generation tasks \cite{dhariwal_diffusion_2021, rombach_high-resolution_2022, mueller-franzes_diffusion_2024}. Their iterative denoising process enables high-fidelity reconstructions by modelling complex data distributions. However, a key limitation of standard diffusion models is their computational cost and speed, as they often require hundreds of iterative steps during inference \cite{rombach_high-resolution_2022, ho_denoising_2020}. This makes them impractical for many real-world applications, due to prolonged inference times. To address this, InDI (Inversion by Direct Iteration) \cite{delbracio_inversion_2023} was proposed as an efficient alternative for conditional image enhancement tasks. InDI reduces the required steps to a fraction by replacing the stochastic reverse diffusion process with a deterministic direct iteration approach. This approach achieves results comparable to traditional diffusion models with significantly fewer computational resources. However, so far, InDI has only been applied to 2D images and non-medical datasets. While classical diffusion models have shown promise in 3D medical image enhancement, their inherent computational cost remains a significant bottleneck for clinical adoption. Generalising efficient 2D frameworks to complex 3D volumetric data and inverse problems such as sparse-view CBCT introduces considerable technical challenges \cite{he_blaze3dm_2024, li_two-and--half_2024, lee_improving_2023, chung_solving_2023}. Thus, our work introduces MInDI-3D, a novel extension of InDI to 3D, which represents a key contribution for enabling efficient high-fidelity, volumetric medical image reconstruction. This gap in the literature motivates our work, which extends InDI to 3D and evaluates its performance in the context of sparse-view CBCT artefact removal.

Our study compares the performance of MInDI-3D with state of the art models. We evaluate the impact of varying training data set sizes and different number of sparse-projections (25 and 50 projections, out of 400 projections in total for the test dataset) on the performance of these approaches. Extensive validation is conducted on test datasets acquired by a different scanner. The perception-distortion trade-off describes the inherent balance in image restoration tasks between achieving high perceptual quality (how "realistic" an image appears to a human observer) and minimising distortion (pixel-level deviations from the original) \cite{blau_perception-distortion_2018}. InDI enables control over this trade-off without retraining: increasing the amount of sampling steps, InDI can trade distortion for better perception reducing the problem of regression to the mean by adding realistic features \cite{delbracio_inversion_2023}. We explore this perception-distortion trade-off. 

\vspace{0.1cm} 

\vspace{0.1cm}
\noindent Our main contributions are summarised as follows:
\begin{itemize}
    \item We introduce MInDI-3D, the first iterative diffusion model adapted for fully volumetric sparse-view CBCT, which provides a unique trade-off between quantitative fidelity and perceptual quality.
    \item We present and publicly release a large-scale pseudo-CBCT dataset of 16,182 volumes to address data scarcity and provide a benchmark for future research in 3D medical image restoration.
    \item We provide a comprehensive validation demonstrating that MInDI-3D enables radiation reduction while maintaining clinically sufficient image quality, confirmed through rigorous quantitative, generalisation, and task-based clinical evaluations by 11 clinicians.
\end{itemize}
\vspace{0.1cm}

\subsection{Related Work}

Extensive research has been conducted on characterising and mitigating artefacts that degrade image quality in CT and CBCT reconstruction \cite{schulze_artefacts_2011, boas_ct_2012}. In recent years, deep learning models have successfully been shown to reduce artefacts in both 3D and 4D (time-resolved) CBCT \cite{amirian_artifact_2024}, offering promising solutions for enhancing sparse-view CBCT image quality (e.g. \cite{amirian_mitigation_2023} for mitigation of motion artefacts). While numerous studies have explored artefact removal in sparse-view CBCT using deep learning, the majority of these approaches have focused on non-generative methods, often employing 2D approaches at times with spatial awareness to reduce computational complexity \cite{wang_improving_2023, hu_prior_2022, jiang_prior_2021}. This spatial compromise creates an opportunity for fully 3D approaches, that by design optimise for inter-slice consistency.

Generative deep learning models in 3D have gained attention in the field of medical imaging for tasks such as unconditional image generation, image-to-image translation (e.g., MRI-to-CT), and image enhancement. Unconditional image generation has been proposed as a privacy‑preserving tool to augment small medical image datasets \cite{khader_denoising_2023}. Three main architecture types have been used in 3D unconditional medical image generation: GANs \cite{kim_3d_2023, liu_inflating_2023}, VAEs \cite{volokitin_modelling_2020, kapoor_multiscale_2023} and diffusion models \cite{friedrich_wdm_2025, khader_denoising_2023}. These developments in unconditional generation have naturally extended to conditional tasks requiring paired data. Image-to-image translation using generative models in 3D has shown impressive results for medical images \cite{pan_synthetic_2024, dorjsembe_conditional_2024, wang_3d_2024, poonkodi_3d-medtrancsgan_2023}. Several studies were conducted using GAN-based approaches, while more recently, researchers have used diffusion and latent diffusion models for medical image to image tasks \cite{friedrich_deep_2025}. For medical image enhancement, generative approaches have seen growing interest, though these approaches remain constrained by computational and practical challenges. While GAN-based approaches dominated early work \cite{xue_cg-3dsrgan_2023, wang_3d_2024, zeng_3d_2022}, recent efforts have shifted toward diffusion-based approaches. 

Our work is most closely related to diffusion-based models used for sparse-view reconstruction, which often leverage multiple 2D diffusion models. For instance, DiffusionMBIR \cite{chung_solving_2023} proposes an effective method for 3D reconstruction by augmenting a pre-trained 2D diffusion model prior with a model-based Total Variation (TV) prior in the z-direction. This TV prior enforces coherence between slices, and the entire process is integrated into an iterative reconstruction framework that includes measurement consistency. Building on the concept of using a 2D diffusion model, the Two-and-a-half-D Score Matching (TOSM) model \cite{li_two-and--half_2024} utilizes a 2.5D fusion technique that combines scores from three orthogonal planes (sagittal, coronal, and transversal) derived from a single pre-trained 2D model to approximate a full 3D score. Similarly, Two Perpendicular 2D Diffusion Models (TPDM) \cite{lee_improving_2023} models the 3D data distribution as a product of two perpendicular 2D plane distributions, performing posterior-based sampling alternatively in each direction. Blaze3DM \cite{he_blaze3dm_2024} integrates a triplane neural field representation with a diffusion model. This approach first constructs compact, data-dependent triplane embeddings from the 3D volumes and then trains a diffusion model on the distribution of these efficient embeddings, significantly reducing computational load. 
DiffusionMBIR, TOSM, TPDM and Blaze3DM were evaluated on the AAPM Low Dose CT Grand Challenge dataset \cite{mccollough_tu-fg-207a-04_2016}, using 9 pseudo CBCT volumes for training and 1 for testing. In contrast, Diffusion Posterior Sampling (DPS) \cite{li_ct_2024} trained on the larger CT Lymph Nodes Dataset \cite{roth_new_2014} while comparing different diffusion-based methods. 

These varied strategies highlight a broader trend towards computationally efficient pseudo-3D solutions \cite{friedrich_deep_2025}. The computational burdens of diffusion-based models has motivated research into efficient diffusion-based implementations that can be practically applied in a clinical setting where reconstruction speed is essential. In this context, InDI is particularly promising, as it requires only a fraction of the sampling steps compared to other diffusion-based models for the conditional setting \cite{delbracio_inversion_2023}.

\section{Materials \& Methods }
\label{method}

\subsection{Datasets}
\label{method:data}

CT-RATE is a public dataset \cite{hamamci_foundation_2024} that includes 25,692 non-contrast chest CT volumes, expanded to 50,188 through various reconstructions, from 21,304 unique patients \autoref{tab:datasets}. From this dataset, we use a subset of 3,612 patients. Volumes were of size $512 \times 512$ voxels in the transverse plane and on average 309 slices along the z-axis and an average spacing of $0.72 \times 0.72 \times 1$ mm on the x, y and z-axis. We used the CT-Rate dataset to generate a pseudo-CBCT training dataset. We forward-projected the CT volumes using a CBCT geometry to obtain CBCT projections (see section \ref{method:recon}), which can then be reconstructed by a CBCT reconstruction algorithm to mimic the CBCT acquisition. Our pseudo-CBCT dataset -- including projection images, sparse-view reconstructions (with 25, 50, and 100 projections), and corresponding ground truth volumes -- is publicly available on Zenodo\footnote{\url{https://zenodo.org/records/XXXXXXXX}}. 

We used a real-world CBCT dataset for testing \autoref{tab:datasets}. This dataset was obtained in a Varian sponsored HyperSight imaging study (acquired on Varian Halcyon linear accelerators). We refer to this dataset as HyperSight. It comprises images from 16 cancer patients including five with abdominal cancer, five with breast cancer, and six with lung cancer, for whom permission to use their data has been obtained. 

\begin{table*}[tbp]
\centering
\begin{tabular}{l|ccccc}
\toprule
\textbf{Dataset} & \textbf{\# Volumes} & \textbf{\# Patients} & \textbf{Anatomic Region} & \textbf{Data Type} & \textbf{Scanner} \\
\midrule
CT-RATE & 16182 & 3612 & chest & pseudo-CBCT & Siemens SOMATOM  \\
HyperSight & 16 & 16 & abdomen, breast, lung  & CBCT & Varian Halcyon \\
\bottomrule
\end{tabular}
\caption{Dataset characteristics showing the pseudo-CBCT training dataset (with both volumes reconstructed with 491 and 697 projections) derived from 8091 CT chest scans (CT-Rate) enabling robust training and 16 real CBCT scans (HyperSight) validating clinical utility across multiple anatomic sites.}
\label{tab:datasets}
\end{table*}

\subsection{CBCT reconstruction and simulation}
\label{method:recon}

Reconstructing 3D CBCT volumes from 2D projections can be achieved through analytical and iterative approaches. The Feldkamp-Davis-Kress (FDK) \cite{feldkamp_practical_1984} algorithm, an analytical method, provides a fast and reliable approximation of the inverse Radon transform, establishing itself as a widely used baseline for 3D CBCT reconstruction. While FDK excels in computational efficiency, iterative reconstruction techniques -- such as the Simultaneous Algebraic Reconstruction Technique (SART) \cite{andersen_simultaneous_1984} -- leverage statistical models and iterative optimisation to improve image quality, particularly in sparse-view or low-dose scenarios. However, their high computational demands often render analytical methods like FDK more practical for routine clinical applications. Our implementation employs FDK with the Ram-Lak filter \cite{ramachandran_three-dimensional_1971} to correct radial sampling non-uniformity, a method commonly termed filtered back-projection (FBP).

While the real-world dataset was acquired using a full-fan, half-trajectory geometry, the pseudo-CBCT was processed with a half-fan, full-trajectory scanning geometry. The full-trajectory configuration involves a 360\textdegree\ rotation, while the half-trajectory rotates  210\textdegree. Half-fan mode allows for a larger field of view by offsetting the detector laterally by 175 mm and using the entire detector for half the field of view. To mitigate artefacts from data redundancy in the overlapping regions of the half-fan geometry, half-fan weighting was applied. The effective area of the real-world detector is $86 \times 43$ cm ($3072 \times 384$ pixels). All projections were generated with a source-to-imager distance (SID) of 1540 mm and a source-to-axis distance (SAD) of 1000 mm. 

For the pseudo-CBCT generation, CT volumes were forward-projected to simulate both full-view and sparse-view acquisitions. Projection parameters -- detector size ($366 \times 160$ pixels), pixel resolution ($1.176 \times 2.688$ mm in axial and longitudinal directions, respectively), and projection counts (491 and 697 for full-view) -- were aligned with a real-world half fan scan protocol from a Varian Halcyon machine. Reconstructions were performed using the FBP method, while varying the number of projections (full, 25, or 50). In sparse-view cases, projections were selected to uniform angular spacing, minimising clustering artefacts and ensuring optimal sampling coverage. The reconstructed volumes have a height, width and depth of $256 \times 256 \times 64$ voxels and a spacing of $2 \times 2 \times 3$ mm. We chose this volume size and spacing to balance memory constraints in our 3D deep learning pipelines with anatomical coverage.

\subsection{Deep Learning Methods}
\label{method:restore}

\begin{figure*}[ht]
   \begin{center}
   \includegraphics[width=\textwidth]{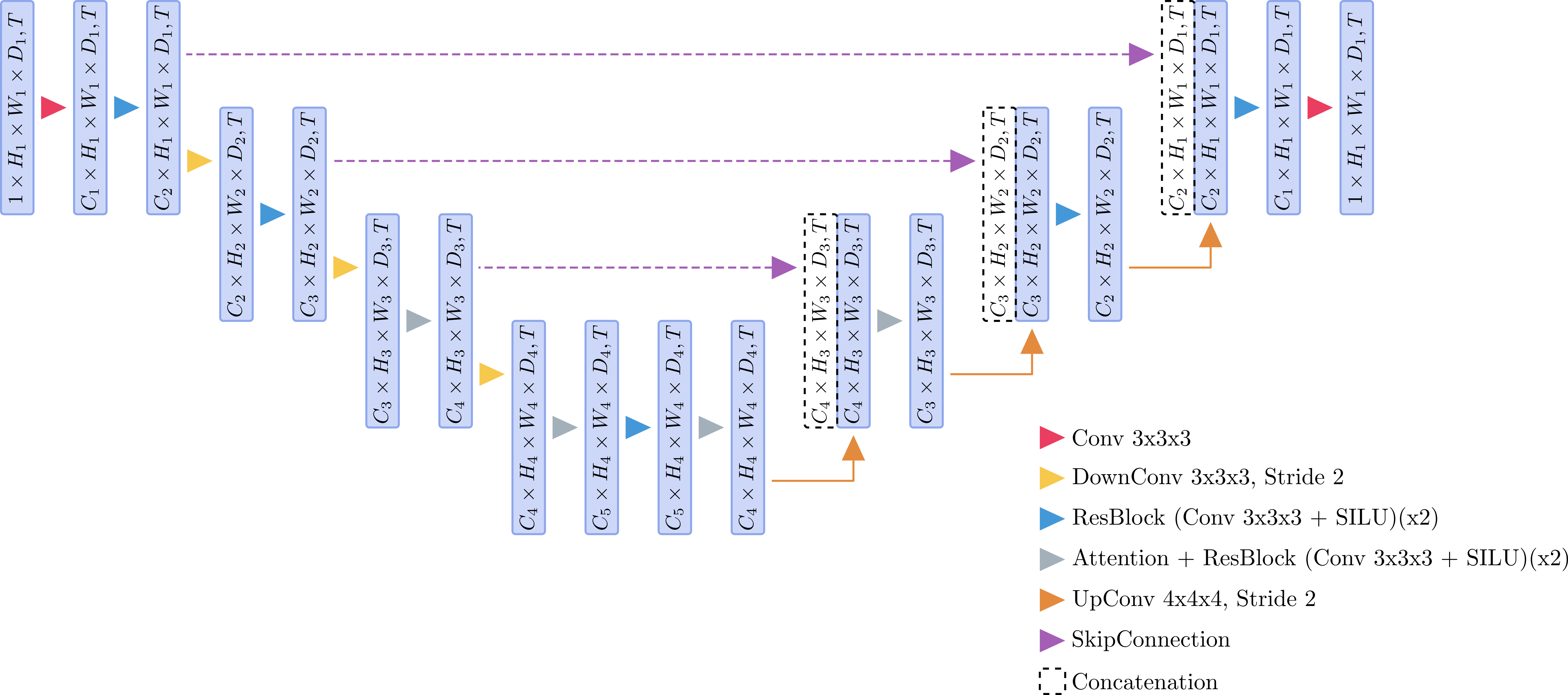}
    \caption{U-Net architecture with 4 hierarchical levels, showing layer-specific dimensionality ($C \times H \times W \times D$), where C is the number of channels, H is height, W is width, and D is depth (all in voxels), and time-embedding (T). SiLU (Sigmoid Linear Unit) activations introduce non-linearity.}
   \label{fig_unet}
   \end{center}
\end{figure*}

This section presents the core methodology for correcting sparse-view artefacts in CBCT images using deep learning. First we present the architecture of our backbone U-Net \cite{ronneberger_u-net_2015, cicek_3d_2016} (see Figure \ref{fig_unet}) and then proceed to the training and inference of MInDI-3D. Unlike many 3D based methods that resort to latent space or explicit spatial compression techniques like wavelets to mitigate memory challenges in volumetric data, our MInDI-3D operates directly in the 3D voxel space to preserve anatomical detail and reduce complexity.

\subsubsection{3D U-Net backbone}
\label{method:restore:unet}

\textbf{Encoder Blocks}:  The encoder comprises four hierarchical stages. Each stage contains two residual submodules followed by downsampling. The first stage contains an additional input layer (kernel: \(3 \times 3 \times 3\), stride: 1). The residual submodules process the volume as follows:  (1) batch normalisation \cite{ioffe_batch_2015} (BN), (2) SiLU activation \cite{hendrycks_gaussian_2023}, and (3) a 3D convolution (kernel: \(3 \times 3 \times 3\), stride: 1). The input to the residual submodule is then added to the output. After the residual blocks, a strided convolution (kernel: \(3 \times 3 \times 3\), stride: 2) downsamples the feature map by a factor of 2. A skip connection adds the stage's output as input to the decoder at the same hierarchical level. Channel dimensions double at each stage, progressing from 32 to 512.  

\textbf{Decoder Blocks}:  The decoder mirrors the encoder, restoring spatial resolution through four stages. Each stage begins by concatenating the skip connection and the output from the lower stage and processing it with a residual submodule described above. The output is then upsampled with a transposed 3D convolution (kernel: \(4 \times 4 \times 4\), stride: 2). Finally, on the last stage, an additional 3D convolutional layer (kernel: \(3 \times 3 \times 3\), stride: 1) is employed. Channel dimensions halve at each stage, reversing the encoder’s progression (512 to 32).  

\textbf{Attention Mechanism}: Convolutional attention applies the Scaled Dot-Product Attention \cite{vaswani_attention_2017} to a convolutional layer following \cite{esser_taming_2021}. The convolutional attention mechanism is integrated into the deepest two encoder and decoder layers and is described subsequently. Input features first undergo group normalisation, followed by a $1 \times 1 \times 1$ convolutions that project the normalised features into query, key, and value tensors. Attention weights are computed via scaled dot-product interactions across all spatial positions in the feature maps, enabling each voxel to dynamically aggregate information from the entire input domain. This global interaction is made tractable by applying the mechanism exclusively at deeper network stages, where hierarchical downsampling has reduced spatial dimensions.

\subsubsection{Inversion by Direct Iteration (InDI)}
\label{method:restore:indi}

InDI is a supervised image restoration method that avoids the "regression to the mean" effect, which can lead to over-correction of outputs toward the average of the training data. By gradually enhancing image quality in incremental steps, InDI produces more realistic and detailed images \cite{delbracio_inversion_2023}. Unlike generative denoising diffusion models, InDI defines the restoration process directly from low-quality to high-quality image, and uses a convex combination of the input/target image as intermediate steps.

\textbf{InDI forward degradation process}: The InDI forward degradation process is defined as follows:
\begin{equation}x_t = (1 - t)x + ty, \quad \text{with} \quad t \in [0, 1].\label{eq_forward_degradation}\end{equation}

\( x_t \) is an intermediate-degraded image between the low-quality input \( y \) (at \( t = 1 \)) and the high-quality target \( x \) (at \( t = 0 \)). The process starts from a clean image at  \( t = 0 \) and degrades it to a noisy image at \( t = 1 \). The iterative restoration process then gradually improves the image quality by moving from \( t = 1 \) to \( t = 0 \) in small steps. 

\textbf{Iterative Restoration Process}: The restoration phase inverts the forward process by iteratively predicting ``cleaner'' images while progressing backward from \( t=1 \) to \( t=0 \). 
\begin{equation}\hat{x}_{t - \frac{1}{N}} = \frac{1}{N \cdot t} \mathcal{F}_\theta\left(\hat{x}_t, t\right) + \left(1 - \frac{1}{N \cdot t}\right) \hat{x}_t
\label{eq_recursive_update}\end{equation}

Equation \eqref{eq_recursive_update} is a recursive update rule from the InDI framework, designed to refine a prediction iteratively. The left-hand side, \(\hat{x}_{t-\frac{1}{N}}\), represents the next predicted time step, with \(N\) representing the number of steps. The right-hand side combines two terms: \(\frac{1}{N \cdot t} \mathcal{F}_\theta\left(\hat{x}_t, t\right)\), which introduces a time-aware backbone model $\mathcal{F}_\theta$. This backbone model predicts the clean image from any time step/ degradation level. \(\left(1 - \frac{1}{N \cdot t}\right) \hat{x}_t\) accumulates the current estimate. As time progresses, the influence of the forward model diminishes, giving more weight to the accumulated estimate, ensuring stability. 

We incorporate a time-embedding into the U-Net backbone of the MInDI-3D model. This time-embedding allows the model to understand the progression from the low-quality image to the high-quality image, effectively encoding the temporal distance between them and enabling an iterative restoration process. We use a sinusoidal time embedding proposed by \cite{ho_denoising_2020} with 1024 channels.

\subsection{Training}
\label{method:training}

Training is conducted on an NVIDIA H200 GPU with 140 GB of VRAM, using the Adam optimiser \cite{kingma_adam_2017} (learning rate 0.0001) and mean absolute error (MAE) (cf. \ref{method:metrics}) as the loss function. To improve convergence, we employ a learning rate scheduler (epoch step size 10, decay factor  $\delta$ = 0.95), a batch size of 4, and gradient accumulation every two steps. Models are trained for 500 epochs, taking approximately 57 hours when using a dataset with 320 subjects for training and 64 subjects for testing. The model with 3200 subjects took 216 hours to train for 180 epochs and was stopped thereafter due to time constraints (412 subjects were used for validation). We optimised the learning rate, learning rate scheduler, batch size, U-Net depth, size and attention layers for optimal performance. Input images were normalised by linearly mapping HU values from -1500 to 1000 onto a range from -1 to 1, without clipping.

\subsection{Metrics \& task-based Evaluation }
\label{method:metrics}

In our experiments, we evaluate numerical distortion performance using several quantitative metrics that measure the point-wise voxel distance between pairs of images $(x,x')$:
\begin{itemize}
\item Mean absolute error $\mathrm{MAE}=\frac{1}{N} \sum_{i=1}^{N}  | x_i - x'_i |$, where \(N\) is the total number of images, \(x_i\) denotes the ground truth voxel value, and \(x'_i\) represents the predicted value;

\item Structural Similarity Index Measure (SSIM)~\cite{wang_image_2004};

\item Peak Signal-to-Noise Ratio \\ $\mathrm{PSNR} = 10 \log_{10} \left( \frac{\mathrm{MAX}^2}{\frac{1}{N} \sum_{i=1}^{N} (x_i - y_i)^2} \right)$, where $\mathrm{MAX}$ is the maximum possible pixel or voxel value;

\item Dice Similarity Coefficient (DICE), which measures the spatial overlap between two segmented volumes, defined as $\mathrm{Dice}(A, B) = \frac{2 |A \cap B|}{|A| + |B|}$, where $A$ and $B$ denote the sets of voxels in the two segmentations. A Dice score of 1 indicates perfect overlap, while 0 indicates no overlap.
\end{itemize}
%
%
%
%
%
%

In \autoref{tab:reconstruction_comparison},  \autoref{tab:InDI_sparse_level}, and \autoref{tab:InDI_ablation}, the standard deviation is shown after the mean of the metric.
All distortion metrics are calculated in Hounsfield units (HU) from pairs of uncorrected or corrected volumes and their corresponding ground truth counterparts. SSIM quantifies structural similarity within spatially correlated 2D/3D regions. Flattening masked data into 1D arrays destroys these spatial relationships, rendering SSIM invalid for masked vectors. MAE and PSNR on the other hand measure pixel/voxel-wise errors, making them suitable for computation on flattened data. They are calculated exclusively for the body, with the air around the body masked out. The masks were generated by first applying Otsu's thresholding \cite{otsu_threshold_1979} method, which automatically determines an optimal threshold value to separate the foreground (typically the region of interest) from the background based on the image histogram. This binary segmentation was then refined using morphological operations. Dilation was used to close small gaps and connect nearby regions, while erosion helped remove small noise and further define the boundaries of the segmented structures. These metrics are referred to as masked. We calculate the PSNR value using 2000 HU as $\mathrm{MAX}$ value, corresponding to a range from -1000 to 1000 HU.

As perception metrics, we use the Fréchet Distance (FD) \cite{stein_exposing_2023} to measure the distance between the distribution of the ground truth compared to the predicted image features. The image features are extracted using the pre-trained model DINOv2 \cite{oquab_dinov2_2024}, which as feature extractor has shown to most closely align with human perception \cite{stein_exposing_2023}. We process the volumes along the axial plane and use the centre slice (2D image) as input for the DINOv2 encoder. 

As task-based evaluation, we use TotalSegmentator~\cite{wasserthal_totalsegmentator_2023} to segment the heart, left lung, right lung, ribs, and vertebrae. We chose TotalSegmentator as a segmentation tool due to its robust segmentation capabilities.

         
\section{Results}
\label{results}
\subsection{Quantitative Evaluation}
The proposed MInDI-3D model was compared with three state of the art diffusion-based models that include spatial information from 3 dimensions: TPDM \cite{lee_improving_2023}, TOSM \cite{li_two-and--half_2024}, DPS \cite{li_ct_2024} and Blaze3DM \cite{he_blaze3dm_2024}. We did not include DiffusionMBIR \cite{chung_solving_2023} as their implementation is focused on fewer projections. We compare the reported metrics with our own results, all of which were evaluated on comparable pseudo-CBCT datasets.
As shown in Table \ref{tab:reconstruction_comparison}, our proposed MInDI-3D model demonstrates strong reconstruction quality compared to current state-of-the-art methods, even when using a more challenging sparse-view scenario with only 25 projections. On the in-domain CT-RATE validation set, MInDI-3D achieves a full 3D PSNR of 36.81 dB and an SSIM of 0.95. This performance is highly competitive with methods like Blaze3DM, which reports a 38.39 dB PSNR from 36 projections using an averaged 2D metric. Most notably, our model achieves a PSNR improvement of +20.20 dB and an SSIM improvement of +0.77 over the FBP baseline, the highest relative gains among all compared methods. On the out-of-domain HyperSight test set, the performance decreases to 29.30 dB PSNR and 0.86 SSIM, demonstrating a sensitivity to domain shift. 

\begin{table*}[tbp]
\centering
\begin{tabular}{p{2.2cm}|c|c|c|p{1.8cm}|p{1.8cm}}
\toprule
\textbf{Method} & \textbf{Projections} & \textbf{PSNR} & \textbf{SSIM} & \textbf{PSNR Improvement} & \textbf{SSIM Improvement} \\
\midrule
\multicolumn{6}{c}{\textbf{Related Work}} \\
\addlinespace
TPDM \cite{lee_improving_2023} & 36 & 38.25 (2D) & 0.949\textsuperscript{*} (2D) & -- & -- \\
TOSM \cite{li_two-and--half_2024} & 29 & 38.21\textsuperscript{*} (2D) & 0.936\textsuperscript{*} (2D) & +15.56\textsuperscript{*} & +0.636\textsuperscript{*} \\
DPS \cite{li_ct_2024} & 30 & 31.29 (2D) & 0.847  (2D) & +12.98  & +0.617 \\
Blaze3DM \cite{he_blaze3dm_2024} & 36 & 38.39 & 0.951\textsuperscript{*} & -- & -- \\
\midrule
\multicolumn{6}{c}{\textbf{Our Work: MInDI-3D}} \\
\addlinespace
Validation set (CT-RATE)        & 25 & 36.81 (3D) & 0.95 (3D) & \textbf{+20.20} & \textbf{+0.77} \\
Test set (HyperSight)           & 25 & 29.30 (3D) & 0.86 (3D) & +10.00 & +0.54 \\

\midrule
\multicolumn{6}{c}{\textbf{Our Work: MInDI-3D (body masked)}} \\
\addlinespace
Validation set (CT-RATE) & 25 & 32.50 (3D, masked) & 0.95 (3D) & +13.45 & \textbf{+0.77} \\
Test set (HyperSight) & 25 & 29.59 (3D, masked) & 0.86 (3D) & + 7.78 & +0.54 \\

\bottomrule
\end{tabular}
\caption{We compare our best results in terms of reconstruction quality with related work. Improvements are computed relative to the analytical FBP reconstruction. \textsuperscript{*} Indicates an average over the three planes while \textbf{bold} indicates the best improvement.}
\label{tab:reconstruction_comparison}
\end{table*}

\begin{figure*}[ht]
   \begin{center}
   \includegraphics[width=1.0\linewidth]{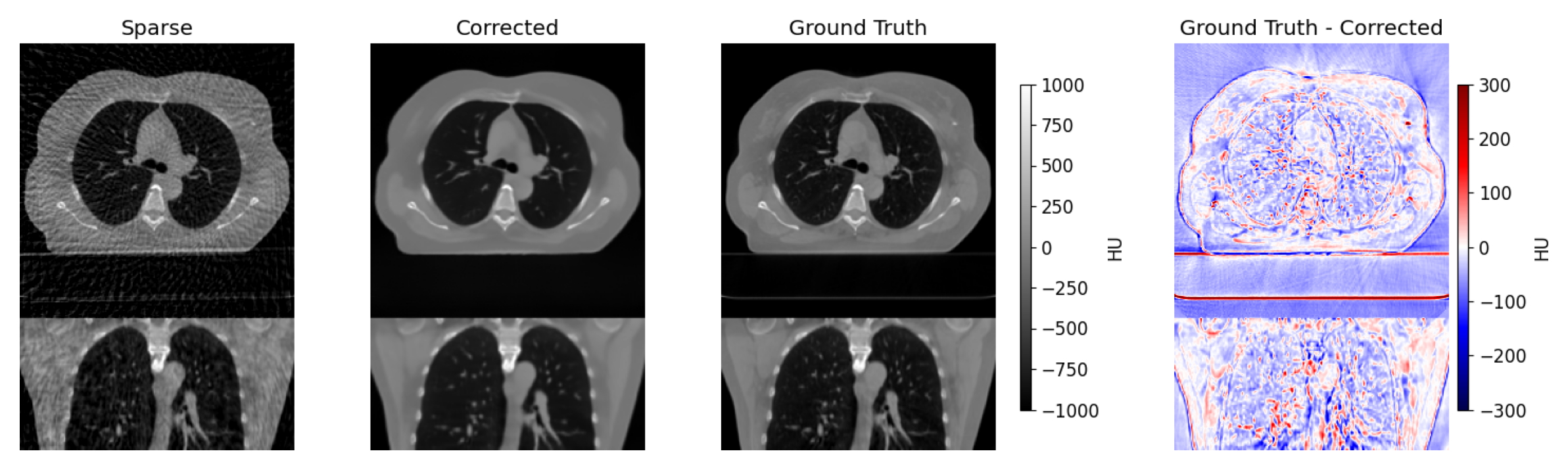}
   \caption{CBCT images (axial and coronal views) of a breast cancer patient (HyperSight dataset), from left to right showing the sparse volume (50 projections), corrected volume using the MInDI-3D model, ground truth volume and a difference plot (ground truth - corrected volume).}
   \label{fig:InDI_image}
   \end{center}
\end{figure*}

To assess robustness to various levels of sparse-view inputs, we trained MInDI-3D with varying projection levels (25, 50, 100) and evaluated on the HyperSight dataset (\autoref{tab:InDI_sparse_level}). The image reconstructed with the smallest number of projections (sparse 25) achieved the largest relative improvement($\Delta_{\mathrm{PSNR}}=+7.78$ dB), compared to the ground truth, while models trained with 100 projections showed the best absolute result ($\mathrm{PSNR}=35.32$ dB).

\begin{table*}[tbp]
\centering
\begin{tabular}{l|c c c}
\toprule
\textbf{Projections} & \textbf{MAE masked $\downarrow$} & \textbf{PSNR masked (dB) $\uparrow$} & \textbf{SSIM $\uparrow$} \\
\midrule

\multicolumn{4}{c}{\textbf{Uncorrected}} \\
\addlinespace
25                         & 125.29 $\pm$ 16.24 & 21.81 $\pm$ 1.15 & 0.32 $\pm$ 0.01 \\
50                         & 65.31 $\pm$ 8.56 & 27.45 $\pm$ 1.18 & 0.47 $\pm$ 0.01 \\
100                        & 27.84 $\pm$ 3.58 & 34.70 $\pm$ 1.21 & 0.70 $\pm$ 0.02 \\

\midrule
\multicolumn{4}{c}{\textbf{MInDI-3D}} \\
\addlinespace
25                         & 48.03 $\pm$ 6.08 & 29.59 $\pm$ 1.00 & 0.86 $\pm$ 0.02 \\
50                         & 30.55 $\pm$ 4.44 & 33.61 $\pm$ 1.16 & 0.91 $\pm$ 0.01 \\
100                        & 24.62 $\pm$ 3.21 & 35.32 $\pm$ 0.94 & 0.93 $\pm$ 0.01 \\

\bottomrule
\end{tabular}
\caption{
MInDI-3D’s performance (2 step) across sparsity levels (25-100 projections) on the HyperSight dataset (MAE, PSNR, SSIM vs. ground truth (mean $\pm$ standard deviation)), where even 25-projection reconstructions achieve 62\% lower MAE than uncorrected scans (48.02 vs. 125.29), validating its potential to enable ultra-low-dose CBCT.}
\label{tab:InDI_sparse_level}
\end{table*}

To quantitatively evaluate anatomical consistency, we performed a task-based analysis using automated segmentation. We applied TotalSegmentator to reconstructions generated by MInDI-3D across a range of iterative refinement steps (1–30). The results demonstrate that segmentation accuracy remains stable, indicating that additional refinement steps do not degrade anatomical integrity. Critical structures like lungs (DICE=0.96–0.99), vertebrae (DICE=0.95) and heart (DICE=0.91–0.92) are preserved and consistency is retained regardless of step count. 

We present an ablation study on the performance of three MInDI-3D models, trained with no, 1, or 2 attention blocks (\autoref{tab:InDI_ablation}). Adding two attention blocks yielded $\Delta_{\mathrm{MAE}}=-5.03$, $\Delta_{\mathrm{PSNR}}=+2.02$ dB and $\Delta_{\mathrm{SSIM}}=+0.01$, validating their importance for capturing global dependencies. Additionally, we analysed the impact of training dataset size on the MInDI-3D model using 64, 320, and 3200 subjects (\autoref{tab:InDI_ablation}). Increasing the training data size improved all metrics, with the 3200-subject model achieving the best metrics, i.e.
 $\Delta_{\mathrm{MAE}}=-11.47$, $\Delta_{\mathrm{PSNR}}=+3.72$ dB and $\Delta_{\mathrm{SSIM}}=+0.03$
 (compared to the 64-subject model), demonstrating the importance of training dataset size for deep-learning based artefact reduction in medical imaging.

\begin{table*}[tbp]
\centering
\begin{tabular}{p{3cm}|c c c}
\toprule
\textbf{Configuration} & \textbf{MAE masked $\downarrow$} & \textbf{PSNR masked (dB) $\uparrow$} & \textbf{SSIM $\uparrow$} \\
\midrule

\multicolumn{4}{c}{\textbf{Uncorrected}} \\
\addlinespace
         & 134.04 $\pm$ 11.02 & 21.15 $\pm$ 0.69 & 0.29 $\pm$ 0.02 \\

\midrule
\multicolumn{4}{c}{\textbf{MInDI-3D: Dataset Size Ablation}} \\
\addlinespace
64 subjects                  & 29.93 $\pm$ 7.15 & 33.53 $\pm$ 1.47 & 0.94 $\pm$ 0.03 \\
320 subjects                & 21.10 $\pm$ 3.36 & 36.08 $\pm$ 1.23 & 0.96 $\pm$ 0.01 \\
3200 subjects              & 18.46 $\pm$ 1.82 & 37.25 $\pm$ 0.84 & 0.97 $\pm$ 0.01 \\

\midrule
\multicolumn{4}{c}{\textbf{MInDI-3D: Attention Block Ablation}} \\
\addlinespace
no attention blocks                 & 26.13 $\pm$ 7.92 & 34.06 $\pm$ 1.53 & 0.96 $\pm$ 0.02 \\
1 attention blocks                 & 22.09 $\pm$ 4.14 & 35.71 $\pm$ 1.40 & 0.97 $\pm$ 0.01 \\
2 attention blocks                  & 21.10 $\pm$ 3.36 & 36.08 $\pm$ 1.23 & 0.97 $\pm$ 0.01 \\

\bottomrule
\end{tabular}
\caption{Ablation study of MInDI-3D (sparse 50, 1 step) performance on the CT-RATE dataset, evaluating (1) training data scalability (64-3200 subjects) and (2) attention block design (0-2 blocks). Larger datasets reduce reconstruction error (3200 subjects: MAE 18.46 vs. 29.93 for 64 subjects), while two attention blocks optimise long-range dependency modeling ($\Delta_{\mathrm{PSNR}}=+2.02$ dB vs. no attention blocks). Metrics averaged over test volumes versus ground truth.}
\label{tab:InDI_ablation}
\end{table*}

\begin{figure}[ht]
   \begin{center}
   \includegraphics[width=1.0\linewidth]{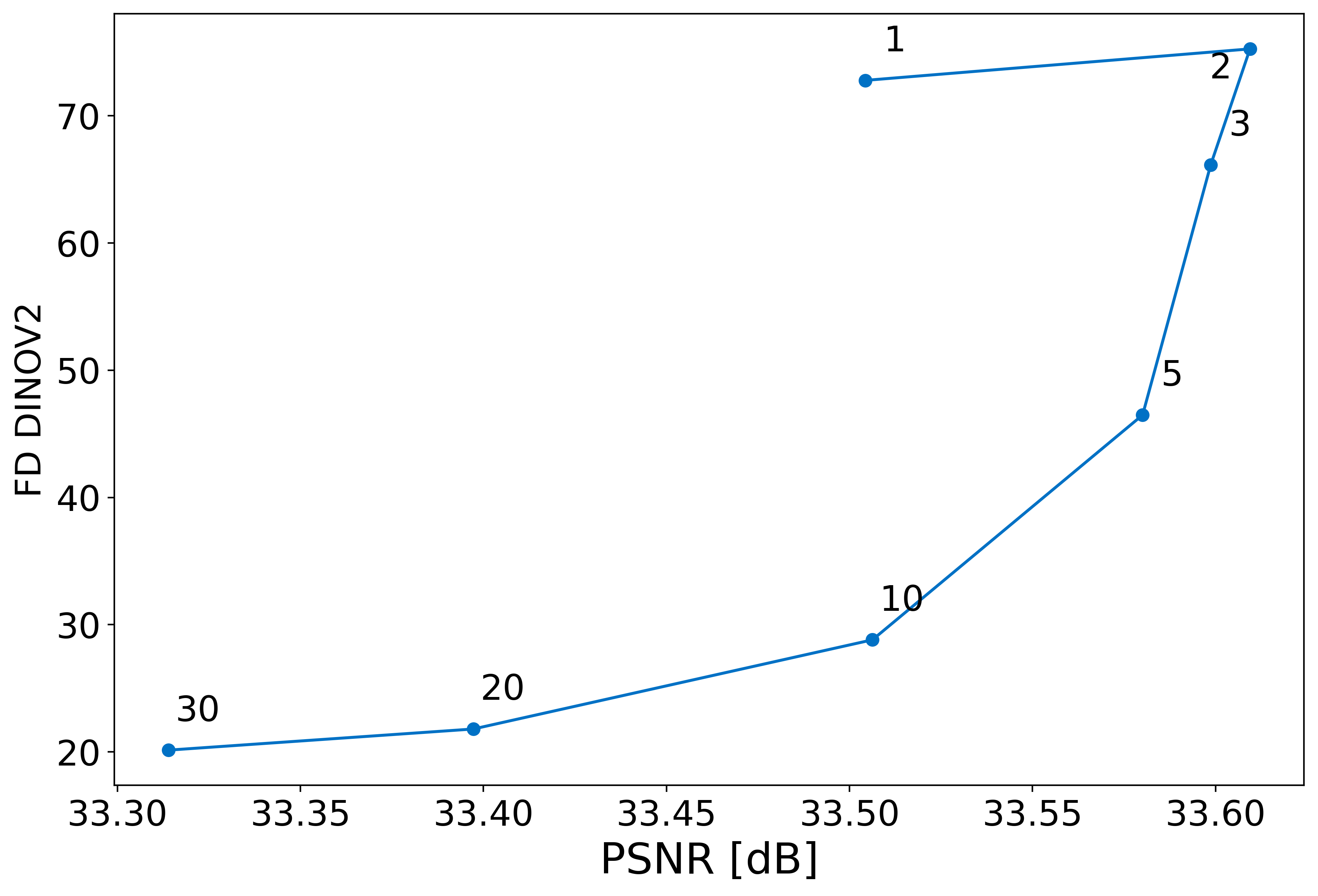}
   \caption{Perception-distortion trade-off in progressive sampling of MInDI-3D on the test set HyperSight with 50 projections. The lineplot compares fidelity (PSNR) and perceptual quality (FD DINOv2) across sampling steps (1-10). Sampling with 2-5 steps improves distortion (higher PSNR) compared to 1 step, while further steps enhance realism (lower FD DINOv2) at the expense of fidelity. Adjusting sampling steps enables precise control over realism and fidelity: steps beyond 2 prioritise perceptual quality, but optimal step counts may vary by anatomy.}
   \label{fig:InDI_perception_distortion}
   \end{center}
\end{figure}

\begin{figure*}[ht]
   \begin{center}
   \includegraphics[width=1.0\linewidth]{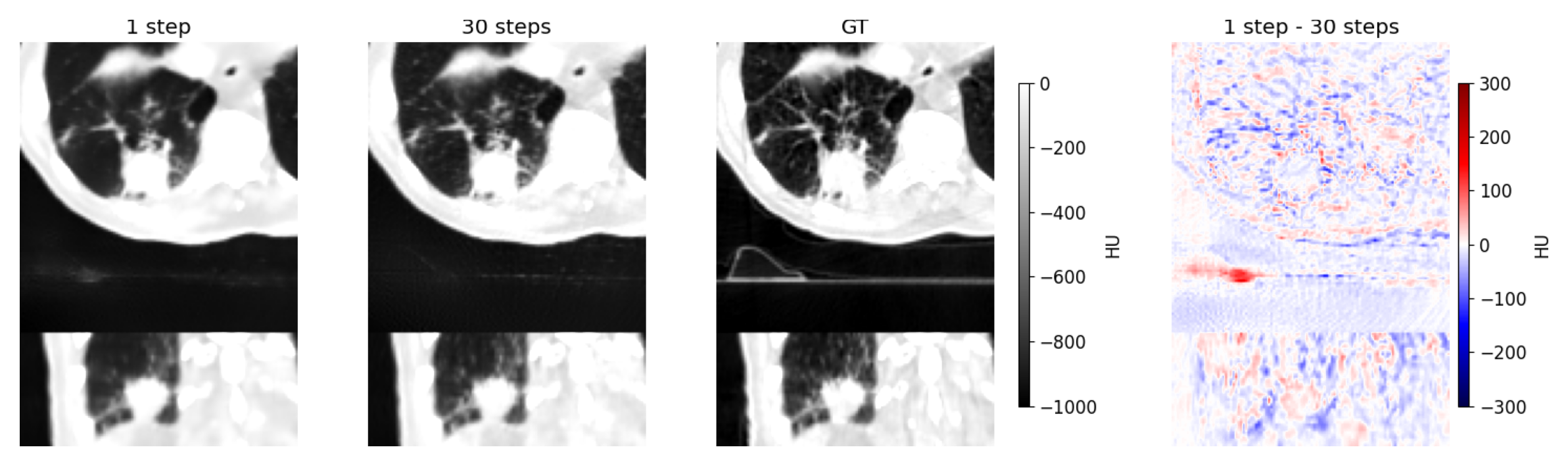}
   \caption{Comparing the MInDI-3D prediction of a lung tumour (lower right lung lobe) from a sparse 50 reconstruction with 1 vs. 30 steps (the ground truth and the difference of step 1 - step 30 as reference). There is an increase of sharpness and detail from step 1 to step 30}
   \label{fig:InDI_perception_distortion_imgs}
   \end{center}
\end{figure*}

We analyse the perception-distortion trade-off in MInDI-3D through progressive sampling (\autoref{fig:InDI_perception_distortion}). A single sampling step yields suboptimal results, failing to optimise either metric. Increasing steps beyond 2 (2-10 steps) trades distortion for realism: PSNR declines modestly (from 33.61 dB to 33.31 dB) while perceptual quality improves (FD DINOv2: from 75.83 to 20.14). This demonstrates that MInDI-3D enables controlled trade-offs between fidelity and realism through step adjustment. Visual examples of this trade-off for a lung tumour are shown in \autoref{fig:InDI_perception_distortion_imgs}, where added steps enhance sharpness and detail. The optimal amount of sampling steps for fidelity, varied across images and anatomic sites. 

\subsection{Clinical Evaluation}
To validate the quantitative results in a clinical setting, a MInDI-3D model -- trained on sparse 50 volumes from 320 subjects -- was tested on the real-world HyperSight dataset. Performance was evaluated based on feedback from 11 clinicians from the Yonsei University Hospital, Seoul, South Korea. The real-world CBCT scans differed from the simulated training dataset, enabling assessment of the models' generalisation capabilities. The primary difference between the training and test datasets was the anatomic site: the training dataset consisted solely of chest CTs, while the test dataset included scans of the abdomen, breast, and lung. Additionally, the geometry used varied, with the training dataset employing half-fan and full-trajectory scans, and the test dataset using full-fan half-trajectory scans.
We provided the clinicians with 16 paired CBCT volumes for review. The sparse volumes were corrected with the MInDI-3D model using 1 inference step and then set side-by side to the full-dose volumes. In every comparison, the tumour was highlighted on the planning CT for reference. The clinicians decided if the corrected sparse-view image was sufficient for any of the following tasks; positioning, contouring and/or dose calculation. The clinicians categorised themselves into the two general categories of radiation oncologist (64\%) and medical physicist (36\%).

For the task of patient positioning, a large part of clinicians agreed that this could be done using the enhanced CBCT volumes for all the anatomical sites investigated (abdomen 96.4\%, lung \& breast 100\%). For the task of dose calculation and contouring, the responses were mixed. The acceptance rates for the AI-enhanced CBCT volumes for dose calculation were 40.0\% for the abdomen, 54.6\% for the breast and 69.7\% for the lung scans. The acceptance rates for contouring were 41.8\%, 80.0\%, 90.9\% for abdomen, breast and lung respectively. Lung scans had the highest acceptance rate, while abdomen scans showed the lowest acceptance rate overall. Overall, the MInDI-3D model demonstrated strong clinical utility for patient positioning across all anatomical sites, with mixed but generally lower acceptance for dose calculation and contouring, particularly in the abdomen, highlighting a need for further refinement in these areas.

\section{Discussion and Outlook}
\label{sec:discussion}
This work introduces MInDI-3D, the first, to our knowledge, adaptation of the InDI framework to 3D and adapted to the medical field. Our findings demonstrate that MInDI-3D not only effectively mitigates sparse-view artefacts, achieving competitive results compared with state of the art models, but also offers unique advantages in terms of a tuneable image quality.

We leverage a large-scale CT dataset via a pseudo-CBCT pipeline, and made the resulting dataset publicly available. This strategy successfully addresses the common limitation of data scarcity in medical imaging, and the observed scaling relationship between dataset size and performance (+3.72 dB PSNR gain) underscores its value. The model's robust generalisation across different anatomies, sparse-levels and unseen acquisition geometries is particularly encouraging. It suggests that MInDI-3D learns fundamental principles of artefact reduction rather than dataset-specific features.

The clinical relevance of MInDI-3D is multifaceted. Task-based evaluations confirm that its iterative refinements preserve crucial anatomical structures, maintaining high segmentation accuracy (e.g., lung DICE $\geq 0.96$) even as perceptual quality is enhanced. This addresses a key concern with generative and deep learning models: ensuring that visual improvements do not compromise diagnostic or treatment-planning information. Direct clinical feedback supports the viability of MInDI-3D for clinical use in specific tasks, such as patient positioning (90-100\%). The clinical tasks of dose calculation and contouring showed more variability between the anatomical sites. The superior acceptance rates for lung scans may reflect both inherent anatomical advantages (high contrast between tumour and surrounding tissue) and domain consistency between training and test data (chest). 

Our results demonstrate that MInDI-3D achieves a state-of-the-art performance, particularly in its ability to reconstruct high-fidelity 3D volumes from as few as 25 projections. The +20.20 dB PSNR and +0.77 SSIM improvement over the analytical FBP reconstruction on our validation set signifies a substantial leap in quality, outperforming the relative gains reported by comparable models. This highlights the model's effectiveness in learning a powerful 3D prior that overcomes the severe ill-posedness of the sparse-view problem. It is also important to note that our evaluation is based on a direct 3D metric, which can be considered a more rigorous assessment than the averaged 2D metrics reported by several other methods. However, the performance drop on the out-of-domain HyperSight test set with a different scanner geometry points to a generalization gap, a critical consideration for clinical translation. This suggests that the learned prior is highly adapted to the training data distribution and that future work should focus on enhancing model robustness across different scanner geometries, anatomies and patient populations.

The perception-distortion trade-off observed with MInDI-3D sampling steps mirrors findings in \cite{delbracio_inversion_2023}. MInDI-3D users can adjust sampling steps to prioritise either quantitative fidelity or perceptual realism, tailoring the output to specific clinical needs. This flexibility is a key advantage, though finding the optimal balance and understanding its clinical implications across diverse scenarios remains an area for further investigation.

Three key considerations arise. First, the pseudo-CBCT simulation, while pragmatic, may not fully replicate real-world scatter and motion artefacts. Second, the perception metric (FD DINOv2) metric, though validated for natural images, requires clinical correlation with radiologist assessments to be validated for a clinical setting, building on \cite{woodland_feature_2024}. Third, while diffusion-based models risk introducing synthetic anatomical features that could mislead clinical interpretation, a critical concern in safety-critical applications like radiotherapy planning, we proactively mitigated this risk through a clinical evaluation. However, future work should rigorously test whether increased sampling steps (3 sampling steps and above) retain anatomical accuracy. This could be done by calculating the treatment dose at different sampling steps to determine whether the dose estimation remains consistent.

Future work should further investigate the trade-off between perceived image quality and anatomical fidelity. Specifically, it is necessary to determine if adding more iteration steps improves clinical usability or if it inadvertently diminishes anatomical accuracy or enhances remaining artefacts. In this context, a systematic study could be conducted on how perception metrics (e.g., FD DINOv2) that were trained on natural images can be utilised to measure perception in 3D in a medical setting. While this study has focused on image enhancement in the pixel space, further research could be conducted in enhancing images in a latent space, which should allow training deep learning models with a higher resolution in 3D.

Our implementation of MInDI-3D establishes conditional generative-based models as viable tools for sparse-view CBCT restoration, achieving clinically acceptable image quality across multiple anatomical sites for certain tasks related to radiation therapy. The framework's generalisation across datasets and scaling with training size highlights the potential of large-scale 3D medical imaging models to advance adaptive radiotherapy. 

\section*{Conflict of Interest}
While some authors are employed by Varian, they declare no conflicts of interest related to this work.

While a direct comparison to other works is challenging due to differing reconstruction geometries, our results demonstrate competitive performance, as shown in \autoref{tab:reconstruction_comparison}. For example, Li et al. \cite{li_two-and--half_2024} reported 2D PSNR improvements of 15.56 and SSIM improvements of 0.636 for a sparse reconstruction with 29 projections. Similarly, Lee et al. \cite{lee_improving_2023} achieved 2D PSNR improvements of  and SSIM of 0.951 with 36 projections. Li et al. \cite{li_ct_2024} achieved a 2D PSNR of 31.29 and SSIM of 0.8471 with 30 projections. It's worth noting that these studies often include background (air) in their error calculations and used smaller datasets (only 10 volumes in \cite{li_two-and--half_2024, lee_improving_2023}) for training and testing. Even with as few as 25 projections, we achieve a PSNR of 36.81 and an SSIM of 0.95 across the entire volume with improvements of 20.20 and 0.77 for PSNR and SSIM respectively.

The perception-distortion trade-off observed with MInDI-3D sampling steps mirrors findings in \cite{delbracio_inversion_2023}. MInDI-3D users can adjust sampling steps to prioritise either quantitative fidelity or perceptual realism, tailoring the output to specific clinical needs. This flexibility is a key advantage, though finding the optimal balance and understanding its clinical implications across diverse scenarios remains an area for further investigation.

Three key considerations arise. First, the pseudo-CBCT simulation, while pragmatic, may not fully replicate real-world scatter and motion artefacts. Second, the perception metric (FD DINOv2) metric, though validated for natural images, requires clinical correlation with radiologist assessments in order to be validated for a clinical setting, building on \cite{woodland_feature_2024}. Third, while diffusion-based models risk introducing synthetic anatomical features that could mislead clinical interpretation, a critical concern in safety-critical applications like radiotherapy planning, we proactively mitigated this risk through a clinical evaluation. However, future work should rigorously test whether increased sampling steps (3 sampling steps and above) retain anatomical accuracy. This could be done by calculating the treatment dose at different sampling steps to determine whether the dose estimation remains consistent.

Future work should further investigate the trade-off between perceived image quality and anatomical fidelity. Specifically, it is necessary to determine if adding more iteration steps improves clinical usability or if it inadvertently diminishes anatomical accuracy or enhances remaining artefacts. In this context a systematic study could be conducted on how perception metrics (e.g., FD DINOv2) that were trained on natural images can be utilised to measure perception in 3D in a medical setting. While this study has focused on image enhancement in the pixel space further research could be conducted in enhancing images in a latent space, which should allow to train deep learning models with a higher resolution in 3D.

Our implementation of MInDI-3D establishes conditional generative-based models as viable tools for sparse-view CBCT restoration, achieving clinically acceptable image quality across multiple anatomical sites for certain tasks related to radiation therapy. The framework's generalisation across datasets and scaling with training size highlights the potential of large-scale 3D medical imaging models to advance adaptive radiotherapy. 

\section*{Conflict of Interest}
The authors declare no conflicts of interest related to this work. Some authors are employed by Varian, as indicated in the affiliations. Their involvement in the study was in accordance with standard academic and ethical practices.


\bibliographystyle{IEEEtran}
\bibliography{IEEEabrv, ./references.bib}

\end{document}